\documentclass[final]{l4dc2020}

\usepackage{ifthen}

\newcommand{\fcnet}[6]{

    \begin{tikzpicture}[shorten >=1pt,->,draw=black!50]

        \tikzstyle{neuron}=[circle,fill=nice-blue!50,minimum size=10pt] %

        \def\layersep{#1}
        \def\nodesep{#2}

        \pgfmathsetmacro{\ilw}{#3}
        \pgfmathsetmacro{\nhl}{#4}
        \pgfmathsetmacro{\hlw}{#5}
        \pgfmathsetmacro{\olw}{#6}

        \pgfmathtruncatemacro{\iof}{(\ilw*\nodesep)/2}
        \pgfmathtruncatemacro{\hof}{(\hlw*\nodesep)/2}
        \pgfmathtruncatemacro{\oof}{(\olw*\nodesep)/2}

        \foreach \name / \y in {1,...,\ilw}
            \path[yshift=\iof]
                node[neuron] (I-\name) at (0,-\y * \nodesep) {};

        \foreach \namei / \yi in {1,...,\nhl}
            \foreach \namej / \yj in {1,...,\hlw}
                \path[yshift=\hof]
                    node[neuron] (H-\namei\namej) at (\yi * \layersep,-\yj * \nodesep) {};

        \foreach \name / \y in {1,...,\olw}
            \path[yshift=\oof]
                node[neuron] (O-\name) at (\nhl*\layersep + \layersep,-\y * \nodesep) {};

        \foreach \source in {1,...,\ilw}
            \foreach \dest in {1,...,\hlw}
                \path (I-\source) edge (H-1\dest);

        \ifthenelse{\nhl > 1}{
            \pgfmathtruncatemacro{\nhlm}{\nhl - 1}
            \foreach \layer in {1,...,\nhlm}
                \pgfmathtruncatemacro{\nextlayer}{\layer + 1}
                \foreach \source in {1,...,\hlw}
                    \foreach \dest in {1,...,\hlw}
                        \path (H-\layer\source) edge (H-\nextlayer\dest);
            }{}

        \foreach \source in {1,...,\hlw}
            \foreach \dest in {1,...,\olw}
                \path (H-\nhl\source) edge (O-\dest);

    \end{tikzpicture}
}

\title{Structured Mechanical Models\\~for Robot Learning and Control}
\author{%
  \Name{Jayesh K. Gupta}\thanks{Equal contribution} \Email{jkg@cs.stanford.edu}
  \AND
  \Name{Kunal Menda}\footnotemark[1] \Email{kmenda@stanford.edu}
  \AND
  \Name{Zachary Manchester} \Email{zacmanchester@stanford.edu}
  \AND
  \Name{Mykel J. Kochenderfer} \Email{mykel@stanford.edu}\\
  \addr Stanford University%
}

\usepackage[utf8]{inputenc} %
\usepackage{booktabs}       %
\usepackage{amsfonts,amssymb,mathtools}       %
\usepackage{subcaption}
\usepackage{multirow}

\usepackage{flushend}
\usepackage{algorithm}
\usepackage[noend]{algpseudocode}
\usepackage{cleveref}
\usepackage{acronym}
\usepackage{verbatim}
\usepackage{siunitx}
\crefname{appsec}{Appendix}{Appendices}

\newcommand\bbR{\ensuremath{\mathbb{R}}} %
\newcommand\qdot{\ensuremath{\dot{q}}} %
\newcommand\qddot{\ensuremath{\ddot{q}}} %
\newcommand\M{\ensuremath{\mathbf{M}}} %
\newcommand\C{\ensuremath{\mathbf{C}}} %
\newcommand\B{\ensuremath{\mathbf{B}}} %

\newcommand\D{\ensuremath{\mathcal{D}}}
\usepackage{todonotes}
\usepackage{soul}
\definecolor{smoothgreen}{rgb}{0.7,1,0.7}
\sethlcolor{smoothgreen}

\usepackage[utf8]{inputenc}
\RequirePackage{luatex85}
\usepackage{pgfplots}
\pgfplotsset{compat=newest}
\pgfplotsset{every axis legend/.append style={%
cells={anchor=west}}
}
\usepgfplotslibrary{polar}
\usetikzlibrary{arrows}
\tikzset{>=stealth'}

\definecolor{C1}{rgb}{0.0, 0.447, 0.741}
\definecolor{C1_light}{rgb}{0.0, 0.6032388663967612, 1.0}
\definecolor{C2}{rgb}{0.85, 0.325, 0.098}
\definecolor{C3}{rgb}{0.929, 0.694, 0.125}
\definecolor{C4}{rgb}{0.494, 0.184, 0.556}
\definecolor{C5}{rgb}{0.466, 0.674, 0.188}
\definecolor{C6}{rgb}{0.301, 0.745, 0.933}
\definecolor{C7}{rgb}{0.635, 0.078, 0.184}

\definecolor{nice-red}{HTML}{E41A1C}
\definecolor{nice-orange}{HTML}{FF7F00}
\definecolor{nice-yellow}{HTML}{FFC020}
\definecolor{nice-green}{HTML}{4DAF4A}
\definecolor{nice-blue}{HTML}{377EB8}
\definecolor{nice-nice-red}{HTML}{984EA3}
\usepgfplotslibrary{groupplots}
\usepgfplotslibrary{dateplot}
\usetikzlibrary{shapes.geometric, arrows}

\tikzstyle{startstop} = [rectangle, rounded corners, minimum width=2cm, minimum height=1cm,text centered, draw=black, fill=none]
\tikzstyle{arrow} = [thick,->,>=stealth]

\usetikzlibrary{backgrounds}

\usepackage{ifthen}

\begin{document}

\maketitle
\begin{abstract}
Model-based methods are the dominant paradigm for controlling robotic systems, though their efficacy depends heavily on the accuracy of the model used.
Deep neural networks have been used to learn models of robot dynamics from data, but they suffer from data-inefficiency and the difficulty to incorporate prior knowledge.
We introduce Structured Mechanical Models, a flexible model class for mechanical systems that are data-efficient, easily amenable to prior knowledge, and easily usable with model-based control techniques.
The goal of this work is to demonstrate the benefits of using Structured Mechanical Models in lieu of black-box neural networks when modeling robot dynamics.
We demonstrate that they generalize better from limited data and yield more reliable model-based controllers on a variety of simulated robotic domains.
\end{abstract}


\section{Introduction}
\label{sec:introduction}

Model-based optimal control has been extensively used in robotics. 
However, the efficacy of this approach can depend heavily on the quality of the model used to predict how control inputs affect the dynamics of the robot. 
While system identification methods can be used to improve the accuracy of a model, models derived from first-principles often fail to accurately express complex phenomena such as friction or aerodynamic drag, suffering from model \textit{bias}.
Although there have been proposals to learn \textit{offset functions} that correct for such model bias, they are usually restricted to specific use cases~\citep{ratliff2016doomed,nguyen2010using}.
To avoid such bias, black-box models such as deep neural networks have been proposed as representations for robot dynamics, with their parameters trained using data collected from the robotic system~\citep{chua2018handful, lambert2019low}.

Black-box neural networks (BBNNs) typically take as input the robotic system's state and control input, and predict its state at the next time-step. 
Though expressive enough to model arbitrary phenomena, such a parameterization has shortcomings.
BBNNs require large amounts of training data to generalize well to unseen states, and such data can be expensive to acquire on real robotic systems. 
The data-inefficiency of these models is a result of the fact that they do not easily incorporate prior knowledge, and are not biased toward solutions corresponding to physical systems. 
Furthermore, many robotic control frameworks, such as feedback linearization, leverage structure present in models derived from first principles~\citep{siciliano2016feedbacklin} and cannot be easily used with BBNNs. 

In this work, we present Structured Mechanical Models (SMMs), which are a flexible black-box parameterization of mechanical systems.
Derived from Lagrangian mechanics, SMMs explicitly encode the structure present in the manipulator equation~\citep{murray1994mathematical}. 
SMMs use the Deep Lagrangian Network architecture~\citep{lutter2018delan} to predict the system's Lagrangian, and use standard neural networks to predict input forces and dissipative forces such as friction.
By parameterizing these different components separately, SMMs retain the expressivity of BBNNs but encode a physical prior that improves their data-efficiency. 
Furthermore, SMMs can be used with a wide variety of model-based control frameworks, including feedback-linearization and energy-based control~\citep{lutter2019energy}.

The goal of the paper is to empirically demonstrate that SMMs are a better black-box parameterization for robotic systems than BBNNs.
Given finite training data, we define ``better'' in the following ways:
\begin{itemize}
    \item SMMs have lower generalization error on unseen states,
    \item SMMs yield more reliable model-based controllers,
    \item SMMs are able to incorporate prior knowledge if available,
    \item SMMs are amenable to the full suite of model-based control techniques.
\end{itemize}

We empirically validate the first two claims on a suite of simulated robotic domains, including a Furuta Pendulum, Cartpole, Acrobot, and Double Cartpole. 
We then discuss the remainder of the claims, potential limitations of SMMs, and how those limitations can be overcome.

\section{Black-box parameterizations of robotic systems}
\label{sec:background}

In this section, we discuss different methods for parameterizing the dynamics of a robotic system without assuming any prior knowledge. 

\subsection{Dynamical and Mechanical Systems}
\label{subsec:dynmechsys}
Dynamical systems are systems with state $x \in \bbR^n$ that evolve over time. 
The time-rate-of-change of the state $\dot{x}$ is specified by a function $\dot{x} = f(x,u)$, and $u\in \bbR^m$ is a control input. 
Mechanical systems are a subset of dynamical systems describing the evolution of extended bodies in the physical world, and can describe many robotic systems of interest. 
The state $x$ of a mechanical system is composed of a set of \textit{generalized coordinates} $q \in \bbR^{n_q}$ and their corresponding time-rates-of-change $\qdot \in \bbR^{n_q}$, called the \textit{generalized velocities}. 
Akin to Newton's second law (force equals mass times acceleration), the dynamics of a mechanical system prescribe the \textit{generalized acceleration}, i.e. $\qddot$, as follows~\citep{murray1994mathematical}:
\begin{equation}
    \label{eqn:manip}
    \M(q)\qddot = F(q,\qdot,u) - \C(q,\qdot)\qdot - G(q)
\end{equation}
Here, $\M(q)$ is the body's \textit{mass matrix}. 
The right-hand side of the equation is composed of the forces that act on the system.
The term $\C(q,\qdot)\qdot$ is the \textit{Coriolis force}, and is entirely a function of $\M(q), q$ and $\qdot$.
The terms $G(q)$ and $F(q,\qdot,u)$ represent the \textit{conservative} and \textit{dissipative} forces, respectively. 
The conservative forces $G(q) = \nabla_q V(q)$ are equal to the gradient of the potential energy $V(q)$, and are termed \textit{conservative} forces because they do not change the total energy of the system.
The dissipative forces are forces such as friction and torques from actuators, which do change the total energy of the system. 
For most real-world mechanical systems, we can write the dissipative forces in \textit{control-affine} form, i.e.:
\begin{equation}
	\label{eqn:controlaffine}
	F(q,\qdot,u) = \tilde{F}(q,\qdot) + \B(q)\cdot u
\end{equation}
Here, the control input $u$ is mapped to a force through the \textit{input Jacobian} $\B(q)$, and all other dissipative forces are captured by $\tilde{F}(q,\qdot)$.
Control-affine formulations are convenient for use with a class of control techniques called \textit{feedback linearization}, where $B(q)$ is inverted to yield a linear control law.

Left-multiplying \Cref{eqn:manip} by $\M^{-1}(q)$, we can isolate $\qddot$ and define $\dot{x} = [\qdot,\qddot]$ in terms of $\M(q)$, $V(q)$, and $F(q,\qdot,u)$. 
This system is defined using a \textit{continuous-time} (CT) formulation. We can convert a CT dynamical system into a \textit{discrete-time} (DT) dynamical system ($x_{t+\Delta t} = f^d(x_t,u_t)$) using an integration scheme such as Runge-Kutta~\citep{runge1895}. 

\subsection{Black-box Models}

The dynamics of a mechanical system are often derived from first-principles~\citep{murray1994mathematical,nguyen2011survey}, but such models can often fail to capture phenomena present in the real world, limiting the efficacy of model-based controllers that use them.

Deep neural networks are a flexible function class that can represent models of almost arbitrary phenomena given training data, and have been used extensively in the field of computer vision~\citep{krizhevsky2012imagenet}. 
Neural networks have been used to represent dynamical systems~\citep{funahashi1993approximation, kuschewski1993application, morton2018deep, chua2018handful,PDDM2019}, and these formulations have been used to perform model-based control~\citep{lambert2019low}.
In these formulations, a neural network directly maps the inputs $x_t$ and $u_t$ to the state derivative $\dot{x}_t$ or next state $x_{t+\Delta t}$.
We call this representation a Black-Box Neural Network (BBNN).
Though flexible enough to represent an arbitrary dynamical system, such a representation does not encode the structure shared by dynamics of all mechanical systems, described in \Cref{subsec:dynmechsys}. 
As a result, BBNNs typically require more data to train, and cannot leverage any prior knowledge one might have about the structure of the mechanical system. 
Furthermore, since such models lack the explicit structure of mechanical systems, they cannot be used with control techniques such as feedback linearization~\citep{siciliano2016feedbacklin} or energy-based control~\citep{spong1996energy}.

\subsection{Deep Lagrangian Networks}
To recover the structure present in the manipulator equation, Deep Lagrangian Networks use neural networks to parameterize the mass-matrix $\M_\theta(q)$ and potential energy of $V_\theta(q)$ of the robotic system, where $\theta$ represents the neural network parameters~\citep{lutter2018delan}. 
The Coriolis forces can be computed using $\qdot$ and the Jacobian of the mass-matrix with respect to $q$ \citep{murray1994mathematical, lutter2018delan}. 
Such model parameterizations have been used to model real robot arms~\citep{lutter2018delan} and for energy-based control of real Furuta pendulums~\citep{lutter2019energy}.
In these works, dissipative forces are assumed to be directly measured, and not predicted.
In the next section, we present Structured Mechanical Models (SMMs), which use neural networks to predict dissipative forces as well as Deep Lagrangian Networks, as a general framework for modeling mechanical systems.

\section{Structured Mechanical Models}
\label{sec:smms}

SMMs are models of mechanical systems that separately parameterize the system's mass-matrix, potential energy, and dissipative and input forces.
If we have no prior knowledge of the system, we use Deep Lagrangrian Networks to parameterize the mass-matrix and potential energy of the system, as well as a neural network $F_\theta(q,\qdot,u)$ that predicts the combination of dissipative and input forces. 
This parameterization is as expressive as a BBNN that predicts $\qddot$ as a function of $q$ and $\qdot$ and $u$.
This is because if we learn $M_\theta(q) = \mathbf{I}$ and $V_\theta(q) = 0$, then the SMM predicts $\qddot = F_\theta(q,\qdot,u)$, which is equivalent to a BBNN. 

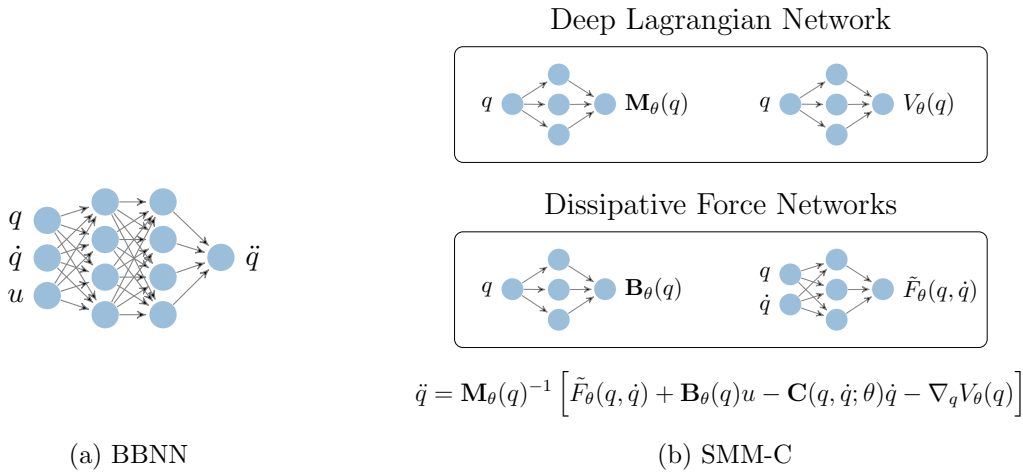
\begin{figure*}
    \centering
   \begin{minipage}[b]{.38\linewidth}
     \centering
    \scalebox{1.0}{\begin{tikzpicture}
    \def\xsep{3.5em}
    \def\bump{0em}

    \def\ls{2em}
    \def\ns{1.3em}
    \node (Nnet) at ($(0,0em)$) {\fcnet{\ls}{\ns}{3}{2}{4}{1}};
    \node[anchor=east] (nqd) at ($(Nnet)+(-\xsep,\bump)$) {$\qdot$};
    \node[anchor=west] (Nqqu) at ($(Nnet)+(\xsep,\bump)$) {$\qddot$};
    \node (nq) at ($(nqd) + (0,1.3em)$) {$q$};
    \node (nu) at ($(nqd) + (0,-1.3em)$) {$u$};

\end{tikzpicture}}
    \vspace{3em}
    \hspace{-1em}
     \subcaption{BBNN}\label{subfig:BBNN}
   \end{minipage}
   \hfill
   \begin{minipage}[b]{.58\linewidth}
     \centering
     \scalebox{0.8}{\begin{tikzpicture}
  \def\xsep{2.5em}
  \def\bump{0em}
  \def\ls{2em}
  \def\ns{1.3em}
  \def\bw{20em}
  \def\bh{5em}
    
  \coordinate (origin) at (0,0);
  \node (potnet) at ($(origin) + (6em,0)$) {\fcnet{2em}{1.3em}{1}{1}{3}{1}};
  \node[anchor=east] (potq) at ($(potnet)+(-\xsep,\bump)$) {$q$};
  \node[anchor=west] (Vq) at ($(potnet)+(\xsep,\bump)$) {$V_\theta(q)$};
  
  \node (mmnet) at ($(origin) - (6em,0)$) {\fcnet{2em}{1.3em}{1}{1}{3}{1}};
  \node[anchor=east] (mmq) at ($(mmnet)+(-\xsep,\bump)$) {$q$};
  \node[anchor=west] (Mq) at ($(mmnet)+(\xsep,\bump)$) {$\M_\theta(q)$};
  
  \node[shape=rectangle, draw=black, rounded corners, minimum height=5em, minimum width=23em, anchor=center] (box1) at ($(origin) + (1em,0)$) {};
  
  \node[anchor=south] at ($(box1.north) + (0,0.15em)$) {\Large Deep Lagrangian Network};
  
  \def\yshift{8em}
  \node (Fnet) at ($(potnet) + (0, -\yshift)$) {\fcnet{2em}{1.3em}{2}{1}{3}{1}};
  \node[anchor=east] (fqq) at ($(Fnet) + (-\xsep,\ns/2)$) {$q$};
  \node (fqqdot) at ($(fqq) + (0,-\ns)$) {$\qdot$};
  \node[anchor=west] (Fqv) at ($(Fnet) + (\xsep, 0)$) {$\tilde{F}_\theta(q,\qdot)$};
  
  \node (Bnet) at ($(mmnet) + (0, -\yshift)$) {\fcnet{2em}{1.3em}{1}{1}{3}{1}};
  \node[anchor=east] (bq) at ($(Bnet) + (-\xsep,0)$) {$q$};
  \node[anchor=west] (Bq) at ($(Bnet) + (\xsep, 0)$) {$\B_\theta(q)$};

  \node[shape=rectangle, draw=black, rounded corners, minimum height=5em, minimum width=23em, anchor=center] (box2) at ($(origin) + (1em,-\yshift)$) {};
  
  \node[anchor=south] at ($(box2.north) + (0,0.15em)$) {\Large Dissipative Force Networks};
  
  \node at ($(box2.south) - (0, 2em)$) {\large $\qddot = \M_\theta(q)^{-1}\left[\tilde{F}_\theta(q,\qdot) + \B_\theta(q)u - \C(q,\qdot;\theta)\qdot - \nabla_q V_\theta(q) \right]$};

\end{tikzpicture}}
     \subcaption{SMM-C}\label{subfig:smm}
   \end{minipage}
    \caption{(a) A black-box model of dynamical system and (b) a Structured Mechanical Model of a control-affine system.}
    \label{fig:bbnnandsmmc}
\end{figure*}

However, almost all mechanical systems are control-affine, i.e. control inputs are mapped to forces via a Jacobian that only depends on $q$. 
Thus, a better black-box parameterization for robotic systems decomposes $F_\theta(q,\qdot,u)$ into $\B_\theta(q)\cdot u$ and $\tilde{F}_\theta(q,\qdot)$ according to \Cref{eqn:controlaffine}, where $\B_\theta(q)$ is a neural network predicting the control-input Jacobian and $\tilde{F}_\theta(q,\qdot)$ is a prediction of dissipative forces not resulting from the input. 
We call this control-affine parameterization SMM-C, and \Cref{fig:bbnnandsmmc} summarizes the networks used in both the BBNN and SMM-C parameterizations.

\subsection{Parameter Optimization}
Given a dataset $\D = \{\ldots, (x_t, u_t, x_{t+\Delta t}), \ldots\}$ of states, inputs, and next states, we find the maximum-likelihood (ML) parameters of an SMM or BBNN by minimizing the mean-squared prediction error of the next state.
That is, we find:
\begin{equation}
    \theta^\text{ML} = \underset{\theta}{\arg\min}\quad \mathbf{E}_{\D} \left[ x_{t+\Delta t} - f_\theta^d (x_t, u_t) \right]
\end{equation}
As previously stated, $f_\theta^d(x_t,u_t)$ is a DT model of a dynamical system. If the model is a CT model, as is the case for SMMs and can be the case for BBNNs, then we apply an integration scheme to convert it to discrete-time.

\section{Experiments}
\label{sec:experiments}

This section justifies the following claims.
Given a finite amount of data gathered from a robotic system:
\begin{itemize}
    \item SMMs typically have lower generalization error on unseen states than BBNNs, and, 
    \item SMMs generally yield more reliable model-based controllers than BBNNs.
\end{itemize}

We demonstrate these claims on four simulated domains---Furuta Pendulum, Acrobot, Cartpole, and Double Cartpole (i.e. Cartpole with two poles attached in series). 
In both experiments, we only make the assumption that the robotic system is control-affine using the SMM-C model parameterization. 

\subsection{Generalization Error on Unseen States}
\label{sec:generr}

In order to test whether SMM-Cs have better generalization error on unseen states, we uniformly sample state, input, and next-state tuples to form training and test datasets.
We then train both an SMM-C model and a BBNN model on the dataset, and measure its generalization error as the mean-squared prediction error on the test set.
We use training sets ranging from 256 to 32768 tuples and a test set with 32768 tuples. 
We run this experiment with 5 random seeds and present standard deviations for performance estimates.

\begin{figure}[t!]
    \centering
    \begin{minipage}[b]{.44\linewidth}
     \centering
    \includegraphics[width=1.0\columnwidth]{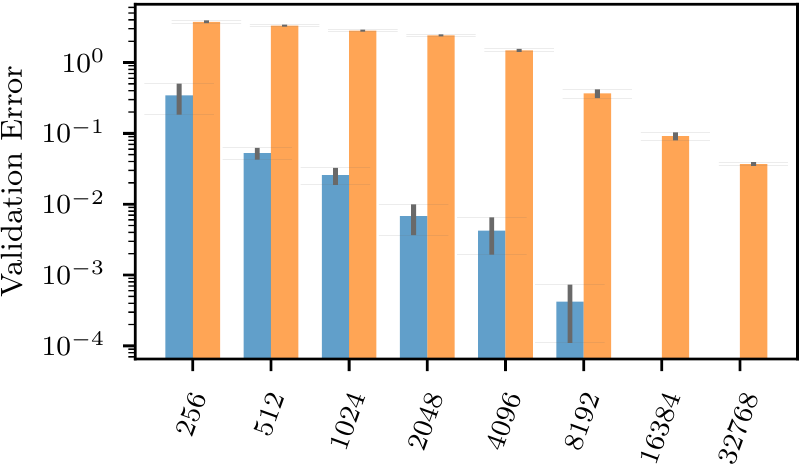}
     \subcaption{Furuta Pendulum}\label{subfig:furuta}
   \end{minipage}
   \hfill
    \begin{minipage}[b]{.44\linewidth}
     \centering
    \includegraphics[width=1.0\columnwidth]{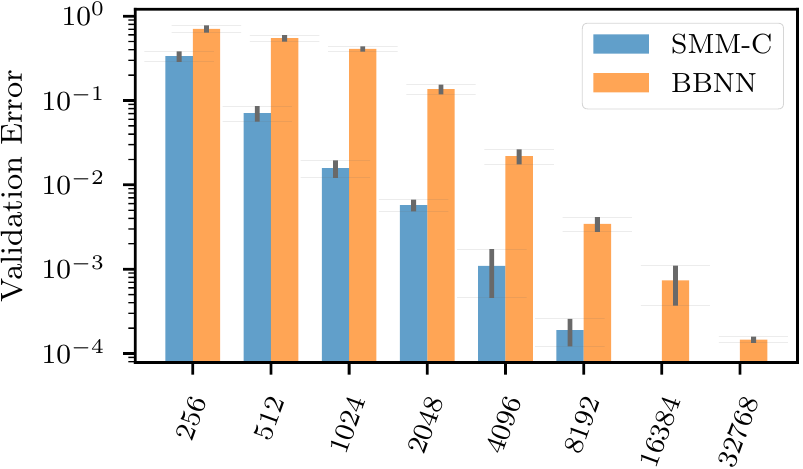}
     \subcaption{Cartpole}\label{subfig:furuta}
   \end{minipage}
   \hfill
   \begin{minipage}[b]{.44\linewidth}
     \centering
    \includegraphics[width=1.0\columnwidth]{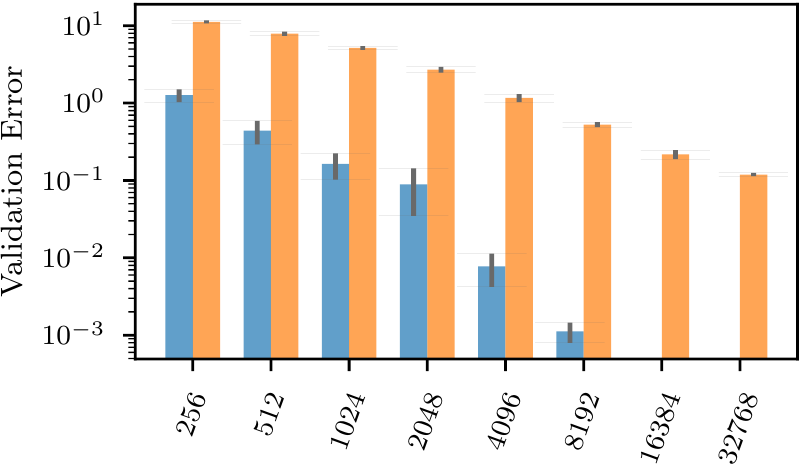}
     \subcaption{Acrobot}\label{subfig:furuta}
   \end{minipage}
   \hfill
   \begin{minipage}[b]{.44\linewidth}
     \centering
    \includegraphics[width=1.0\columnwidth]{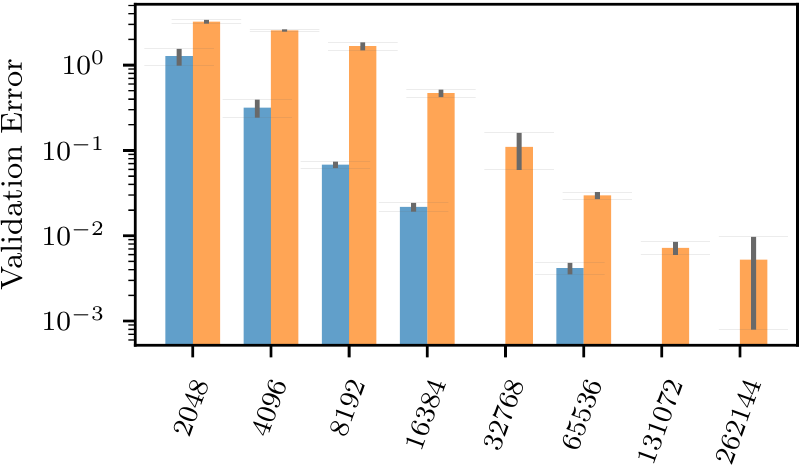}
     \subcaption{Double Cartpole\label{subfig:furuta}}
   \end{minipage}
   
    \caption{Generalization error of SMM-Cs and BBNNs for various training dataset sizes on four robotic domains.}
    \label{fig:generr}
\end{figure}

\textit{Results:} \Cref{fig:generr} summarizes the generalization error of a BBNN and SMM-C parameterization of the dynamics of the four robotic domains, for various dataset sizes. 
We see that for all domains and training set sizes, the generalization error of the SMM-C parameterization is at least an order of magnitude better, and up to three orders of magnitude better than that of the BBNN parameterization. 
The disparity between the models' abilities to generalize from limited data can be explained by the following two facts:
Firstly, BBNNs confound the forces generated by Coriolis effects, potential fields, control inputs, and dissipative sources, unlike SMMs which explicitly decouple these effects, and therefore serve as a better prior for the true model structure.
Secondly, neural networks in SMM-Cs do not have any explicit dependence on the generalized velocities, aside from the function $\tilde{F}(q,\qdot)$ that predicts disspative forces. 
As a result, SMM-Cs generalize far better to states with unseen velocities. 
Next, we show that even if enough data is provided such that an SMM-C and BBNN achieve similar generalization errors, SMM-Cs are more reliable when used for model-based control.

\subsection{Reliability of Model-Based Control}
In this experiment, we attempt to solve Furuta Pendulum, Cartpole and Acrobot swing-up tasks by using the SMM-C and BBNN models trained in \Cref{sec:generr}. 
That is, starting at their respective stable equilibria, a control sequence must be applied to bring the systems to and stabilize them at their respective unstable equilibria.
These tasks are canonical examples of model-based control problems because they are underactuated, i.e. the control input cannot independently actuate the degrees-of-freedom of the robot. 
Though many approaches to model-based control exist, a popular approach to solving underactuated tasks is to first compute a \textit{nominal trajectory} and then follow the nominal trajectory. 

The nominal trajectory $(\bar{x}_{1:T}, \bar{u}_{1:T-1})$ is a series of states and inputs that are both dynamically-feasible according to the model, i.e. $x_{t+1} = f_\theta^d(x_t,u_t)$, and connect the initial and goal configurations with minimum cost. 
It is common to minimize a quadratic cost of the form: 
\begin{equation}
    J(x_{1:T}, u_{1:T-1}) = (x_T - x_g)^\top Q_F (x_T - x_g) + \sum_{t=1}^{T-1} (x_t - x_g)^\top Q (x_t - x_g) + u_t^\top R u_t
\end{equation}
Here, $x_g$ corresponds to the goal-state, and $Q$, $Q_F$, and $R$ are positive definite matrices trading-off the trajectory's efficiency and control effort.
In our experiments, we use Direct-Collocation Trajectory Optimization (DIRCOL)~\citep{kelly2017trajopt,howell2019altro} to find the nominal trajectory by solving a constrained non-linear program.

Once the nominal trajectory is found, it must be followed by the robotic system in a manner that rejects disturbance. 
To do so, a feedback controller of the form  $u_t = \pi(x_t, t) = \bar{u}_t + K_t (\bar{x}_t - x_t)$ is used to compute that input $u_t$ at time $t$. 
The \textit{feedback gains} $K_t$ are computed using a Time-Varying Linear Quadratic Regulator (TVLQR)~\citep{tvlqr}.
The resulting controller is then run on the true system, and the quality of the resulting trajectory is largely dependent on the accuracy of the model used in DIRCOL and TVLQR.
Letting the resulting trajectory be $(x^\pi_{1:T}, u^\pi_{1:T-1})$, we can measure the \textit{cost} of the trajectory as:
\begin{equation}
    J^\pi(x^\pi_{1:T}, u^\pi_{1:T-1}) = \underset{x_t^\pi}{\min}\quad \lVert x_t^\pi - x_g \rVert^2_2
\end{equation}
If a model class is able to consistently generate policies that solve the swing-up task, the resulting cost will be low, and we will consider the model class reliable.

We note that TVLQR synthesizes the gains $K_t$ using cost matrices $Q^\text{\tiny TV}, R^\text{\tiny TV}$, and $Q_F^\text{\tiny TV}$ and the resulting cost $J^\pi$ is sensitive to the choice of these hyperparameters. 
To make a fair comparison between model classes, we perform a grid search over these hyperparameters and report the lowest cost found.

In this experiment, we compare the five SMM-C and BBNN models trained with 8192 samples on each domain.
In addition, we will also compare the five BBNN models trained on Cartpole with 32768 samples, referred to as BBNN-32768, since it has a generalization error comparable to that of the SMM-C models we are using.
By doing so, we hope to explore if the structure present in SMMs aids model-based control beyond merely having low prediction error on unseen states.
We report the mean and standard deviation of $J^\pi$ across the five models.

\def\aa{$(5.24 \pm 3.14) \times 10^{-3}$}
\def\ab{$(6.29 \pm 8.98) \times 10^{0}$}
\def\ac{*}
\def\ad{$(7.86 \pm 0.0) \times 10^{-3}$}

\def\ba{$(1.35 \pm 0.99) \times 10^{-3}$}
\def\bb{$(1.89 \pm 1.14) \times 10^{1}$}
\def\bc{$(1.80 \pm 1.17) \times 10^{1}$}
\def\bd{$(6.40 \pm 0.0) \times 10^{-4}$}

\def\ca{$(1.55 \pm 2.94) \times 10^{-1}$}
\def\cb{$(0.84 \pm 1.03) \times 10^{1}$}
\def\cc{*}
\def\cd{$(1.48 \pm 0.0) \times 10^{-4}$}

\begin{table}[]
\resizebox{\textwidth}{!}{\begin{tabular}{rcccc}
\toprule
                                                                           & {SMM-C}  & {BBNN}   & BBNN-32768 & {True}   \\
\cmidrule(lr){2-5}
Furuta Pendulum & \aa & \ab & \ac                                            & \ad \\
Cartpole        & \ba & \bb & \bc & \bd \\
Acrobot         & \ca & \cb & \cc                                            & \cd \\
\bottomrule
\end{tabular}}
\caption{Performance of learned SMM-C and BBNN models on the Furuta Pendulum, Cartpole and Acrobot swing-up tasks, compared to the true model.\label{tab:swingup}}
\end{table}

\textit{Results:} \Cref{tab:swingup} summarizes the performance of model-based controllers synthesized using BBNNs and SMM-Cs on the two robotic domains tested. 
We see that SMM-Cs yield model-based controllers that achieve far lower costs when attempting to solve swing-up tasks, and is comparable in cost to the true model\footnote{Videos of the attempted swing-ups and source code for the experiments can be found at \url{https://sites.google.com/stanford.edu/smm/}.}.
This observation can largely be explained by the fact that the BBNN models tested have worse generalization error than the SMM-C models tested. 

\def\errS{$(1.9 \pm 0.7) \times 10^{-4}$}
\def\errB{$(1.2 \pm 0.1) \times 10^{-4}$}

\def\JxxS{$(4.1 \pm 6.7) \times 10^{-2}$}
\def\JxuS{$(1.2 \pm 1.7) \times 10^{-4}$}
\def\JxxB{$(0.5 \pm 1.1) \times 10^{-1}$}
\def\JxuB{$(0.5 \pm 2.9) \times 10^{-2}$}

However, we see that even the BBNN-32768 models, which have marginally better generalization errors than the SMM-C models trained on Cartpole (\errB~for BBNN-32768 and \errS~for SMM-C), perform worse than the SMM-C models.
We hypothesize that in order to perform DIRCOL and TVLQR, we not only need accurate predictions of next states given unseen states and inputs, but also require accurate estimates of Jacobians with respect to the states and inputs. 
We find that on the same test-set $\D_\text{test}$ used in \Cref{sec:generr}, the Frobenius-norm error in predictions of $\nabla_x f^d_\theta(x,u)$ is \JxxS~and of $\nabla_u f^d_\theta(x,u)$ is \JxuS~for the SMM-C, but is \JxxB~and \JxuB~respectively for the BBNN-32768 trained on Cartpole, which is consistent with our hypothesis.
Hence, we see evidence that SMM-Cs are a better black-box parameterization for model-based controllers.

\section{Discussion and Conclusion}
\label{sec:conclusion}

In the last section, we empirically validated the first two claims---that SMM-Cs are a better black-box parameterization than BBNNs in terms of generalization to unseen states, and yield more reliable model-based controllers.
Though we used black-box parameterizations that make no assumptions about the system aside from that it is control-affine, SMM-Cs have the additional benefit of being able to incorporate any prior knowledge available to a practitioner.
For example, we may know the system's mass matrix or kinematic structure and can replace the neural network predictor with a model derived from first principles.
Additionally, we may want to model specific dissipative forces, like linear joint friction, instead of a black-box predictor, and so we can replace $\tilde{F}_\theta(q,\qdot)$ with a more structured parameterization.
Lastly, SMM-Cs can be used with the full suite of optimal controllers. 
Since the model is a control-affine system, they can easily be used with feedback-linearization techniques, unlike BBNNs. 
Additionally, since we explicitly predict kinetic and potential energies, we can use energy based-control, which has been demonstrated on a physical Furuta pendulum~\citep{lutter2019energy}. 

There are caveats to our study of SMM-Cs. 
We have only demonstrated SMM-Cs on robotic domains with smooth dynamics, while BBNNs have been demonstrated as useful model parameterizations on simulated contact-rich domains~\citep{nagabandi2018finetuning, chua2018handful, PDDM2019}. 
However, SMMs can be used in contact-rich domains with the use of variational integrators~\citep{manchester2017variational}. 
To do so, kinematic constraints resulting in contact are explicitly satisfied when making state predictions by solving a constrained optimization problem, where contact and non-holonomic constraints are enforced using Lagrange multipliers. 
Furthermore, constraints can be introduced and removed after the model is trained.

BBNNs used in contact-rich domains directly predict next-states~\citep{nagabandi2018finetuning}, as opposed to predicting state derivatives as presented in this work.
The shortcoming of such an approach is that the kinematic constraints that result in contact can only be implicitly satisfied by minimizing prediction loss, with no guarantee that predictions made at test-time will satisfy the constraints. 
Additionally, if a new kinematic constraint is introduced, the BBNN will have to be re-trained. 
These properties warrant the study of SMMs in place of BBNNs for black-box modeling in contact-rich domains, and is an avenue for future work.

Another caveat worth mentioning is training and query time.
We empirically find that it takes on the order of 5$\times$ more CPU time per training epoch to train an SMM-C than to train a BBNN on the same amount of data. 
This is because more computation is required for SMM-C predictions compared to a BBNN. 
Some of this could be mitigated with various numerical and software optimizations.

\acks{We are grateful to Jeannette Bohg for advice. This work is supported in part by DARPA under agreement number D17AP00032. The content is solely the responsibility of the authors and does not necessarily represent the official views of DARPA.}

\bibliography{references}

\end{document}